\def\year{2021}\relax
\documentclass[letterpaper]{article} % DO NOT CHANGE THIS
\usepackage{aaai21}  % DO NOT CHANGE THIS
\usepackage{times}  % DO NOT CHANGE THIS
\usepackage{helvet} % DO NOT CHANGE THIS
\usepackage{courier}  % DO NOT CHANGE THIS
\usepackage[hyphens]{url}  % DO NOT CHANGE THIS
\usepackage{graphicx} % DO NOT CHANGE THIS
\usepackage{natbib}  % DO NOT CHANGE THIS AND DO NOT ADD ANY OPTIONS TO IT
\usepackage{caption} % DO NOT CHANGE THIS AND DO NOT ADD ANY OPTIONS TO IT
\usepackage[colorinlistoftodos,prependcaption,textsize=tiny]{todonotes}
\usepackage{amsmath}
\usepackage{lipsum}                     % Dummytext      
\usepackage{xargs}                      % Use more than one optional parameter in a new commands
\newcommandx{\unsure}[2][1=]{\todo[linecolor=red,backgroundcolor=red!25,bordercolor=red,#1]{#2}}
\newcommandx{\change}[2][1=]{\todo[linecolor=blue,backgroundcolor=blue!25,bordercolor=blue,#1]{#2}}
\newcommandx{\add}[2][1=]{\todo[linecolor=green,backgroundcolor=green!25,bordercolor=green,#1]{#2}}
\newcommandx{\improvement}[2][1=]{\todo[linecolor=Plum,backgroundcolor=Plum!25,bordercolor=Plum,#1]{#2}}
\usepackage{subcaption}
\usepackage[switch]{lineno}

\usepackage{siunitx}

\title{A spatio-temporal LSTM model to forecast across multiple temporal and spatial scales}

\author{
    Yihao Hu,\textsuperscript{\rm 1}
    Fearghal O'Donncha,\textsuperscript{\rm 2} 
    Paulito Palmes,\textsuperscript{\rm 2} \\
    Meredith Burke,\textsuperscript{\rm 3}
    Ramon Filgueira,\textsuperscript{\rm 3} 
    Jon Grant.\textsuperscript{\rm 3}
    \\
}
\affiliations{
    \textsuperscript{\rm 1}University of Notre Dame\\
        \textsuperscript{\rm 2}IBM Research Europe\\
                \textsuperscript{\rm 3}Dalhousie University\\

    yhu5@nd.edu, feardonn@ie.ibm.com, paulpalmes@ie.ibm.com, \\
    meredith.burke@dal.ca, ramon.filgueira@dal.ca, 
    jon.grant@dal.ca
}
\begin{document}

\maketitle
\begin{abstract}
This paper presents a novel spatio-temporal LSTM (SPATIAL) architecture for time series forecasting applied to environmental datasets. The framework was evaluated across multiple sensors and for three different oceanic variables: current speed, temperature, and dissolved oxygen. Network implementation proceeded in two directions that are nominally separated but connected as part of a natural environmental system -- across the spatial (between individual sensors) and temporal components of the sensor data. 
Data from four sensors sampling current speed, and eight measuring both temperature and dissolved oxygen evaluated the framework.
Results were compared against RF and XGB baseline models that learned on the temporal signal of each sensor independently by extracting the date-time features together with the past history of data using sliding window matrix.
Results demonstrated ability to accurately replicate complex signals and provide comparable performance to state-of-the-art benchmarks. Notably, the novel framework provided a simpler pre-processing and training pipeline that handles missing values via a simple masking layer. 
Enabling learning across the spatial and temporal directions, this paper addresses two fundamental challenges of ML applications to environmental science: 1) data sparsity and the challenges and costs of collecting measurements of environmental conditions such as ocean dynamics, and 2) environmental datasets are inherently connected in the spatial and temporal directions while classical ML approaches only consider one of these directions. Furthermore, sharing of parameters across all input steps makes SPATIAL a fast, scalable, and easily-parameterized forecasting   framework.

%%%RESULTS         Bidirectional      Baseline      % reduction
% Oxygen     MAE:    0.5857           0.9201
% Temperature MAE:   0.7834           6.3509
% ADCP       MAE: 

\end{abstract}
\section{Introduction}

% \textcolor{red}{
% Potential names for the bidirectional LSTM.
% \begin{itemize}
%     \item bidirectional LSTM
%     \item Spatial bidirectional LSTM
%     \item spatiotemporal LSTM
%     \item ST-LSTM -- Let's stick with this as naming: ST-LSTM (spatiotemporal-LSTM)
% \end{itemize}
% }
%A~time series is a sequence of data point indexed in time order that are  ubiquitous in scientific, industrial, and economic activity. Time-series analyses extends across several areas of study including forecasting, data mining, or characterising salient explanatory variables. In this paper, we study time series forecasting for environmental applications -- a complex process that requires resolution of multiple physical and biological processes. More specifically, we apply a bidirectional LSTM architecture to forecast ocean- temperature, current speed, and dissolved oxygen. 

The coastal ocean represents perhaps the most complex modeling challenge, intimately connected as it is to the deep ocean and the atmosphere \citep{song1994semi}.
Irregular coastlines, and steep highly variable bottom topography can generate highly complex patterns of flow; wind forcings produce both surface and internal waves and contributes to surface flows directly through wind drift and Ekman transport; tidal forcing in addition to barotropic flow processes induce internal tides; while freshwater discharges add buoyancy fluxes, which further complicate local water motion \citep{odonncha2015characterizing}.
Consequently, the accurate quantification of flow processes in the ocean is a complex, nonlinear task requiring a holistic consideration of dynamics and driving forces in both space and time.
Further, understanding, quantifying, and forecasting ocean dynamics is of major importance for societal and economic factors: we look to the oceans for food, water, and energy, while ocean-atmosphere dynamics are critical to understanding weather and climate.

%While univariate time-series approaches are well studied, most real-world \mbox{problems} are not that simple. The prediction is influenced by multiple variables, and processes at neighbouring locations drive local dynamics and variations. While time \textit{and} space dependent problems are a fundamental part of engineering and mathematics, they are little-studied in machine learning. Classical examples of mathematical descriptions of space-time systems include the Navier-Stokes equation \citep{constantin1988navier} in fluid mechanics, and the convection-diffusion-reaction equation with application in domains that stretch from air pollution \citep{stockie2011mathematics} to economics \citep{barles1998option}. 

Traditionally, forecasting of ocean processes rely on physics-based approaches that resolve a set of governing equations such as Navier-Stokes \citep{constantin1988navier}  or convective-diffusion equations \citep{stockie2011mathematics}.
%Classical approaches for forecasting in environmental application relies primarily on physics-based models that resolve a set of governing equations. 
An extensive body of literature exists on this subject with early investigations on the topic described in \citet{bryan1969climate} and a useful overview provided in \citet{ji2017hydrodynamics}. 
The primary challenge of physics-based approaches are the immense computational expense to deploy at high resolution across broad spatial and temporal scales (generally requiring high performance computing facilities), and the complexity of configuring and parameterizing the model that typically requires an expert user \citep{deyoung2004challenges}.

Machine-learning (ML) approaches to forecast ocean processes are in a nascent stage and have traditionally been limited by the challenge of data sparsity. To circumvent this limitation, recent years have seen interest in using ML to develop low cost approximates or surrogates of physic-based models (e.g. \citet{lary2004using, ashkezari2016oceanic}. In this paradigm, there is the luxury of being able to run the model as many times as necessary to develop a sufficient data set to train the ML model. A key issue with this approach is that one is limited by the performance of the physics-based model; as an example \citet{James2018} have recently observed a factor of 12,000 improvement in the speed of obtaining results comparable to that of the physics model, but because the model itself is utilized to train the deep-learning network, the accuracy and spatial extents are limited to the levels dictated by the traditional approaches. 

Progress in satellite technology generates large volumes of remotely sensed observations providing long-term global measurements at varying spatial and temporal resolution. These datasets provide a natural fit for ML-based approaches and there is a wide body of literature related to mining and forecasting ocean processes from satellite observations. 
%Progress in satellite technology has increased the granularity of measurements that are possible, providing long-term global measurements at varying spatial and temporal resolution enabling significant advances in ML-applications.
Deep Learning (DL) models have been adopted to increase the resolution of satellite imagery through down-scaling techniques \cite{Ducournau_deep_2016}, while a~number of studies have presented data-mining approaches that extract pertinent events from measured data such as detection of harmful algal blooms \cite{Gokaraju_machine_2011}. Some caution is necessary here since satellite estimates of ocean processes are invariably contaminated by external influence on the optical properties \citep{holt2009modelling, smyth2006semianalytical}, while one is naturally limited to sampling surface processes and cannot penetrate the ocean depth. 

This paper investigates a framework that extends model accuracy by explicitly learning across the spatial and temporal components of a disparate -- but related -- time-series signal. We refer to our framework as \textit{Spatial-LSTM} or SPATIAL.

SPATIAL is applied to three real-world datasets, namely, ocean current speeds, temperatures, and dissolved oxygen.
Incorporating both physical and biogeochemical datasets, the performance of the model with different temporal scales and process memory are evaluated:
ocean current patterns are influenced by external physical drivers such as tidal effects and surface winds stress together with density driven flows generated by changes in temperature and salinity;
variations in ocean temperature are explained by multiple exogenous processes such as solar radiation, air temperature, sea bed heat transfer, and external heat flux from rivers and the open ocean \citep{pidgeon2005diurnal};
 oceanic oxygen content, on the other hand is a biogeochemical process influenced by hydrodynamics (horizontal and vertical mixing, residence times, etc.), weather (temperature reducing the solubility of oxygen), and nutrient loads (both anthropogenic and organic enrichment) \citep{caballero2015biogeochemical}.
The combination of nonlinear response and sensitivity to multiple, opaque variables makes the capabilities of DL an appealing solution for these prediction-requirements.

%This study developed a deep neural network prediction framework for ocean current speed, temperature, and dissolved oxygen. 

%Quantifying and forecasting ocean dissolved oxygen concentrations are an important area of study and application due to its ecological importance and sensitivity to climatic perturbations \citep{andrews2017biogeochemical}. 

\section{Related Work}

There is a large number of application papers on using ML for time-series forecasting.
For classical (linear) time series when all relevant information about the next time step is included in the recent past, multilayer perceptron (MLP) is typically sufficient to model the short-term temporal dependence \citep{gers2001lstm}.
Compared to MLPs, convolutional neural networks (CNN) automatically identify and extract features from the input time series.
A~convolution can be seen as a sliding window over the data, similar to the moving average (MA) model but with nonlinear transformation.
Different from MLPs and CNNs, recurrent neural networks (RNN) explicitly allow long-term dependence in the data to persist over time, which clearly is a desirable feature for modeling time series.
A review of many of these approaches for time series-based forecasting is provided in \citet{esling2012time, gamboa2017deep, fawaz2019deep}. 

Forecasting of environmental variables relies heavily on physics-based approaches and a well-established literature exists related to hydro-environmental modeling and applications to numerous case studies. An excellent operational forecasting product are the Copernicus Marine Service ocean model outputs which provide daily forecasts of variables such as current speed, temperature, and dissolved oxygen (DO) at a 10-day forecasting window \citep{Copernicus_2020}.  Data are provided at varying horizontal resolution from approximately \SI{10}{km} globally, to regional models at resolution of up to \SI{3}{km}. These datasets serve as a valuable forecasting tool but are generally too coarse for coastal process studies and for most practical applications (e.g. predicting water quality properties within a bay) are rescaled to higher resolution using local regional models (e.g. \citet{gan2005open}. 

Due to the heavy computational overhead of physical models, there is an increasing trend to apply data-driven DL or ML methods to model physical phenomena \cite{bezenac_deep_2017,wiewel_latent_2018}.
A~number of studies have investigated data-driven approaches to provide computationally cheaper surrogate models, applied to such things as wave forecasting \cite{James2018}, viscoelastic earthquake simulation \cite{devries_enabling_2017}, and water-quality investigation \cite{arandia_surrogate_2018}. In all examples, the ML has been used to perform non-linear regression between the inputs and outputs.

An alternative approach aims to embed information from physics or heuristic knowledge within the network. 
Physics-informed DL is a novel approach for resolving information from physics. The philosophy behind it is to approximate the quantity of interest (e.g., governing equation variables) by a deep neural network (DNN) and embed the physical law to regularize the network. To this end, training the network is equivalent to minimization of a well-designed loss function that contains the PDE residuals and initial/boundary conditions \citep{rao2020physics}.

A further stream of related work has been started by \citet{NIPS2018_7892}, who presented a novel approach to approximate the discrete series of layers between the input and output state by acting on the derivative of the hidden units. At each stage, the output of the network is computed using a black-box differential equation solver which evaluates the hidden unit dynamics to determine the solution with the desired accuracy. In effect, the parameters of the hidden unit dynamics are defined as a continuous function, which may provide greater memory efficiency and balancing of model cost against problem complexity. The approach aims to achieve comparable performance to existing state-of-the-art with far fewer parameters, and suggests potential advantages for time series modeling.
For follow-up work in this stream, see \citet{grathwohl2018scalable,gulgec2019fd}.

% Physics model-based approaches\ldots \\
% Some related literature on statistical time series forecasting\ldots \\
% Deep Learning\ldots \\
% Dedicated feature engineering to incorporate external information\ldots \\
% Transfer learning\ldots \\
% Physics models + machine learning; surrogate models\ldots \\
% Feature augmentation\ldots \\
% Physics-informed loss function\ldots\\

\begin{figure*}[t]
\centering
    \includegraphics[width=\textwidth]{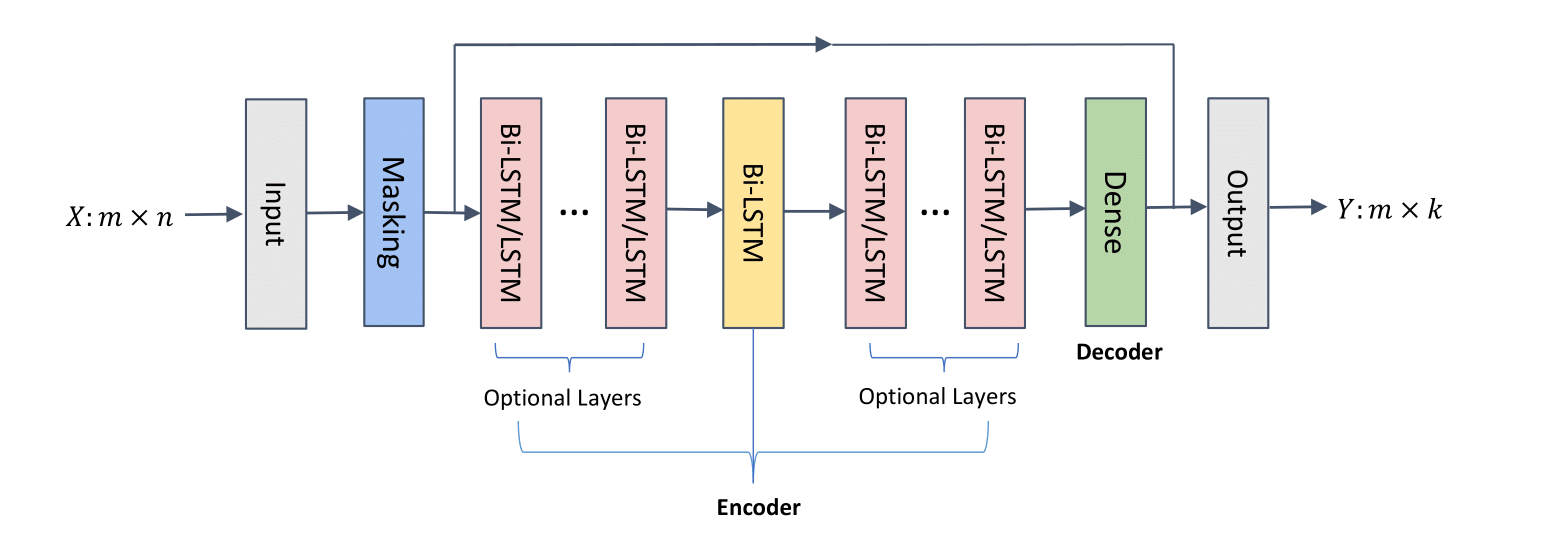}
    \caption{SPATIAL allows multidimensional input, which enables the deep neural network to extract features from different sensors and exploit the information learned to predict the time series for every sensor. That's being said for each sensor, the prediction is not only based on its previous time series, but also the information from other sensors.}
    \label{fig:spatialoverview}
\end{figure*}\vspace{10mm}
\begin{figure*}
\begin{subfigure}[h]{0.45\linewidth}
\includegraphics[width=\linewidth]{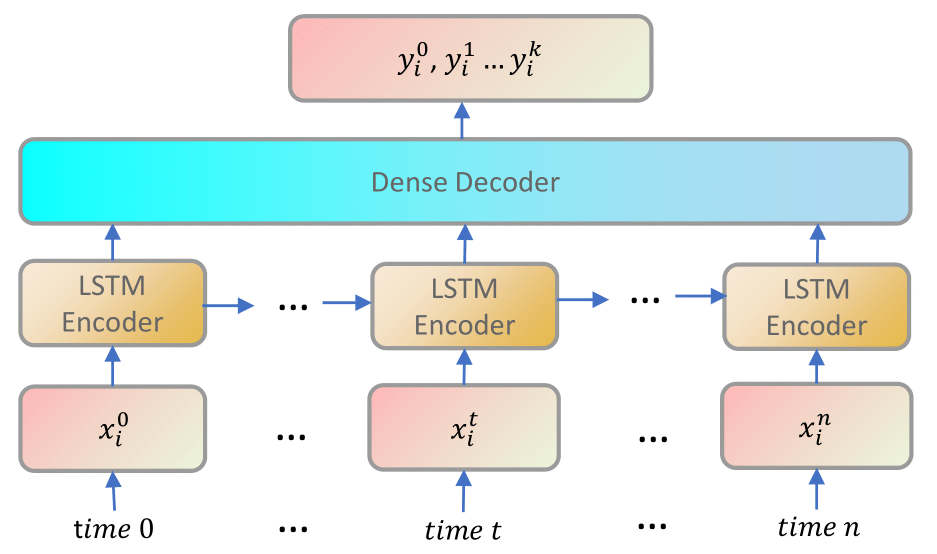}
\caption{Classical LSTM Model for single sensor $i$}
\end{subfigure}
\hfill
\begin{subfigure}[h]{0.45\linewidth}
\includegraphics[width=\linewidth]{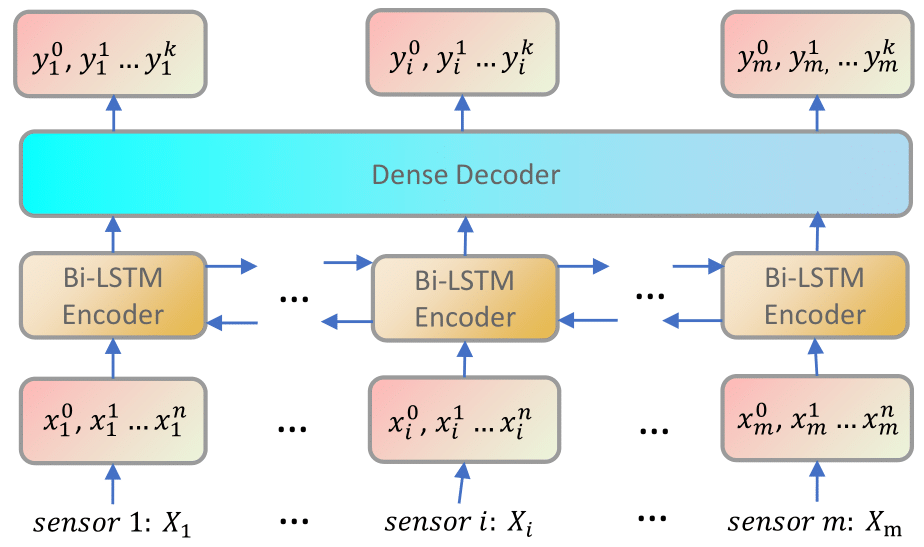}
\caption{Unfolded Orthogonal Bidirectional LSTM model}
\end{subfigure}%
\caption{Left figure (a) presents a schematic of a classical LSTM architecture to predict a single time series signal while (b) presents our SPATIAL implementation in which the spatial (sensor-to-sensor) and temporal patterns are explicitly learned by the network. In this representation, \textbf{m} is the number of sensors, \textbf{n} is the training time period, and \textbf{k} is the time period we wish to predict.}
\label{fig:lstm_architecture}

\end{figure*}

\section{Methodology} 
\label{sec:methods}
SPATIAL is a first attempt at using a bidirectional LSTM model on both the spatial and temporal directions of a time-series signal. More specifically, based on geographical proximity or domain expertise, it is known that the signals that we wish to predict (ocean currents, temperature, and dissolved oxygen) have some dependency on neighbouring signals. We implement a bidirectional LSTM model across the \textit{spatial} direction of pertinent sensors and train the model to learn the spatial and temporal structure. An overview of SPATIAL is provided in Figure~\ref{fig:spatialoverview}, and described in detail in the following section. 

% \subsection{Machine learning automated pipeline leak detection}
% \unsure{for AAAI might not need this}
% Feature engineering and how to automate\ldots\\
% Touch on Lale and Julia pipeline\ldots \\

\subsection{Deep Neural Networks}
\label{subsec:dnn}
%Theories and concepts of bidirectional LSTMs\ldots\\
RNN and its variants are widely studied for time series forecasting \citep{hochreiter1997long}.
 A~fundamental extension of RNNs compared to ANN approaches is parameter sharing across different parts of the model. This has natural applicability to the forecasting of time-series variables with historical dependency. 
 A schematic representation is provided in Figure~\ref{fig:lstm_architecture}(a) illustrating the neural network mapping framework from an input time series, $(x_{1},x_{2}, \cdots, x_{t})$, to output label vector, $ (y_{t+1}, y_{t+2},\cdots, y_{T})$.  In effect, the RNN has two inputs, the present state and the past.
 
 Standard RNN approaches have been shown to fail when lags between response and explanatory variables exceed 5--10 discrete timesteps \cite{gers_lstm_1999}. Repeated applications of the same parameters can give rise to vanishing, or exploding gradients leading to model stagnation or instability.
A number of approaches have been proposed in the literature to address this, with the most popular being LSTM \cite{GoodBengCour16}.

An extension of the parameter sharing enabled by recurrent networks is bidirectional LSTM which processes sequence data in backward and forward directions \citep{schuster1997bidirectional}. This has natural application to sequential or time series data, and has demonstrated improved performance against unidirectional LSTM in areas such as language processing \citep{wang2016image}, and speech recognition \citep{graves2013hybrid}.
While bidirectional LSTM has previously been applied to time series forecasting, this paper is the first that applies across the spatial as well as the temporal components. 
Figure \ref{fig:lstm_architecture}(b) provides a schematic of our SPATIAL implementation. A unidirectional LSTM is applied across the time direction of each sensor, and between individual sensors a series of stacked bidirectional layers support learning between distinct time series. The number of stacked layers is a hyperparameter selected during training. 
The features to the network consist of an  \textbf{m} 
$\times$ 
\textbf{n} array (where \textbf{m} is number of sensors and \textbf{n} is number of time points) and labels consist a corresponding \textbf{m} $\times$ \textbf{k} array (where \textbf{k} is the prediction window).

\subsection{The data}

In this study, SPATIAL is applied to the following datasets: ocean current speed or velocity collected along the Norwegian coast, ocean temperature, and dissolved oxygen sampled from a high-density network of sensors at an aquaculture site in Atlantic Canada. 

Acoustic Doppler Current Profiler (ADCP) is used to measure water current velocities over a depth range, using the Doppler effect of sound waves scattered back from particles within the water column. Four ADCPs were deployed at individual locations along the Norwegian coast. The sensors collect 1-minute interval measurements at vertical resolution or bins of between 5 and \SI{15}{m} over a 1-year period (January 1\textsuperscript{st} -- December 31\textsuperscript{st} 2018). Observations were averaged over the depth to generate a single time series for each ADCP. 

Table~\ref{tab:adcp} provides summary metrics of the ADCP deployments including longitude, latitude, and distance between ADCPs (relative to arbitrarily selected, ADCP 1). The data were downloaded from the OPeNDAP server provisioned by the Norwegian Meteorological Institute \citep{norkyst_opendap}. While these sensors are geographically distant (up to \SI{60}{km} apart), theory imparted from transfer learning \citep{pan2009survey} suggests that knowledge of the dynamics at one location could be used to guide training at others.

\begin{table}
    \begin{center}
        \begin{tabular}{|l | r | c | c | c|} 
        \hline
         \# & ADCP name & Long & Lat & Dist. (km) \\ [0.5ex] 
         \hline\hline
       1 & E39\_A\_Sulafjorden & 6.18 & 62.56  & 00.0 \\ 
        \hline
         2 & E39\_B\_Sulafjorden & 6.35 & 62.67 & 22.4 \\
         \hline
         3 & E39\_D\_Breisundet & 6.09 & 62.60 & 11.1 \\
         \hline
        4 & E39\_F\_Vartalsfjorden & 5.74 & 62.16 & 65.2 \\
        \hline
        \hline
        \end{tabular}
    \end{center}
    \caption{The geographical longitude and latitude coordinates, and reference name (complying with the naming convention used on the OpeNDAP portal \citep{norkyst_opendap} of the four ADCP datasets used in this study, together with the distance between each sensor and the first one listed}
    \label{tab:adcp}
\end{table}

Temperature and dissolved oxygen observations were collected within an aquaculture farm of sea cages cultivating Atlantic salmon. 
Each cage was \SI{100}{m} in circumference, \SI{12}{m} deep, with around 25,000\,fish. The cages were equipped with an array of sensors sampling both temperature and dissolved oxygen. We gathered information from eight sensors distributed across a cage: 4 at \SI{2}{m}, and
4 at \SI{8}{m} (North, South, East, West configuration).

These sensors sample a complex system described by natural conditions (natural ocean variation due to dynamics) and the effects of fish farming (modified temperature stratification patterns, consumption of oxygen by fish, etc.). Furthermore, the geographical proximity of cages exerted an influence: modification of upstream dynamics will be amplified at downstream sensors as effects accumulate (e.g. oxygen depletion) \citep{odonncha2013physical}. 
Tidally driven flows have the most significant effect on dissolved oxygen levels, with its influence a function of cage location within the farm. 
As waters flow from one end of the farm to the other, fish behaviour and physiology, as well as flow restriction from cage infrastructure reduce oxygen levels. This results in significant oxygen variations throughout the farm with high oxygen levels at one end of the farm and low oxygen levels at the other. Since farmers use low oxygen as an indicator as to whether to feed, it is important to implement robust predictive models that reflect spatial variations at high resolutions to enable informed decisions that impact fish growth and welfare \citep{burke_do_2020}. 

The DO and temperature data was collected over a four-month period from August 16\textsuperscript{th} -- December 21\textsuperscript{st} 2018 over relatively small geographical extents, characterised by significant dynamics and aquaculture-industry activities. Additional details on the data collection and application to directing aquaculture operations are provided in \citet{burke_do_2020}.

\subsection{Data Preprocessing}

Anomalous data and gaps in the sensor data are common in ocean monitoring. Table \ref{tab:summarystat} presents summary statistics including information on the frequency of occurrence of missing data. The \textit{BMean} column quantifies the mean length of contiguous blocks of missing data for each sensor dataset (averaged across all sensors). The ADCP data in particular reports large data gaps which complicates learning and requires bespoke data cleaning routines to adjust to the characteristics of the models.

Each individual sensor data consists of a one-dimensional time series at 1-minute interval. We resampled each dataset to 30-minute intervals to avoid spurious fluctuations while maintaining the fidelity of the data. Data processing proceeded according to the requirements of the particular ML pipeline. 

SPATIAL has the simplest data preparation requirement with each time series loaded into a single array of $[N_{sensors}, N_{timepoints}]$  for each of the three variables. 

\begin{table}[htbp]
    \begin{center}
        \begin{tabular}{|r | c | c | c | c| c|} 
        \hline
         Dataset     & Mean       & $\sigma$     & Skewness   & Kurtosis & BMean \\ [0.2ex] 
         \hline\hline
             ADCP    & 20.22      &  7.54        &   2.51     &  9.11    & 216.61   \\ 
        \hline
             Temp    & 12.21      &  0.85       &  -0.30     &  -1.34    & 4.36   \\
         \hline
               DO    & 08.50     &   0.55       &  -0.13     & -0.66   &  4.44  \\
         \hline
        \hline
        \end{tabular}
    \end{center}
    \caption{Summary statistics of the three datasets.}
    \label{tab:summarystat}
\end{table}

The AutoAI and AMLP pipelines follow the tabular ML workflow of data preparation. We used the TSML~\cite{tsml2020} which is an open-source package to automatically convert the time series 
data into matrix form by \emph{sliding window}. Each row represents the autoregressive features together with the date-time features (year, month, day hour, day of week, etc), time-aligned with the corresponding label at the desired prediction window. We setup the problem as a 24-hour ahead prediction using half hour stride with 6-hour autoregressive window size predetermined based on the initial experiments.

For the AutoAI and AMLP approaches, missing data were flagged and any rows that contained missing data were dropped in the sliding window matrix generated. SPATIAL  on the other hand simply applied a masking layer that acted on the data flagged as missing and automatically ignored them during learning.

The complete source code of SPATIAL has been released publicly under Apache  license  at \url{https://github.com/IBM/spatial-lstm}, whereby further details of the implementation can be studied in detail. The ADCP data we used is freely available and the open source code contains scripts to download and preprocess it. We currently do not have permission to release the oxygen and temperature data as it pertains to commercial aquaculture operations.

\begin{figure}[t]
\centering
    \includegraphics[width=0.49\textwidth]{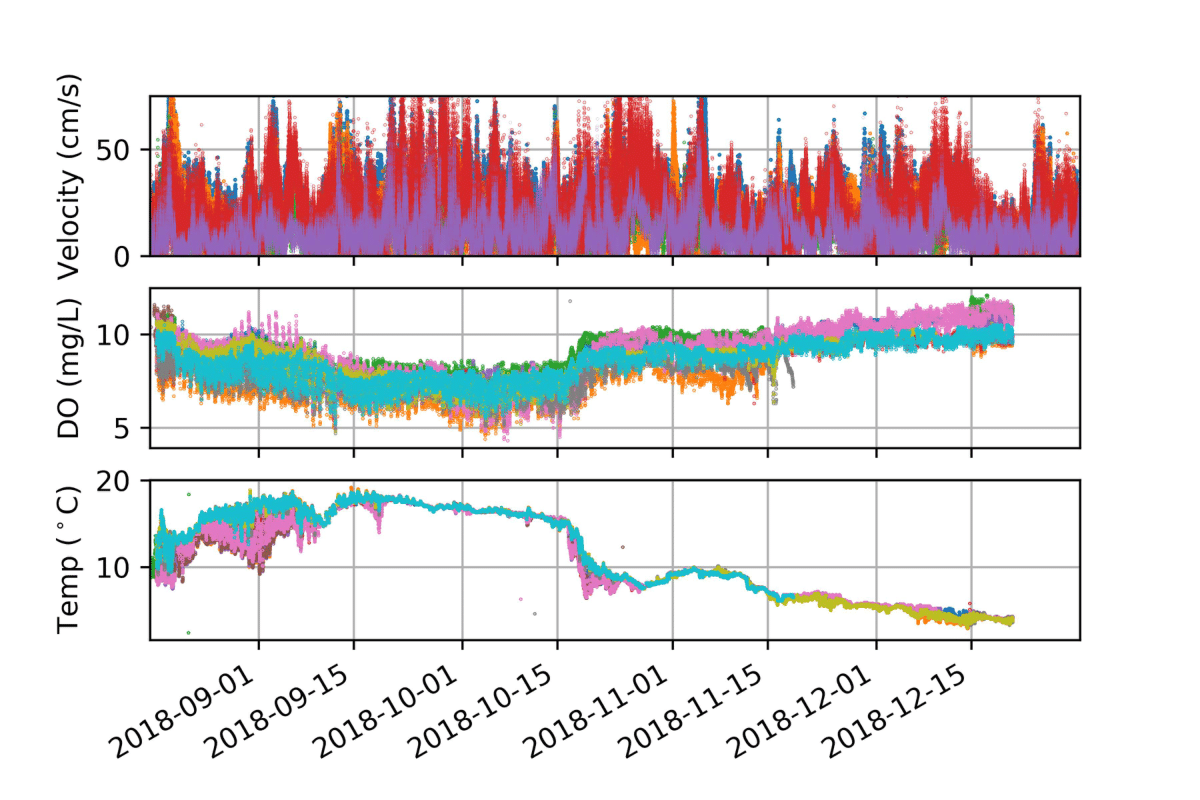}
    \caption{Time series plot of the three datasets. Top figure represents measurements of ocean current speed from four locations (summarised in Table XXX), while middle and bottom figure denotes dissolved oxygen and temperature observations respectively at eight locations.}
\label{fig:allsensordata}
\end{figure} 

\section{Results and Discussion}
\label{sec:res}

To evaluate the relative performance of SPATIAL with respect to existing approaches, we included two optimal baseline models based on the AutoML~\cite{drori2018} technology:
\begin{itemize}
    \item IBM AutoAI \citep{IBM_autoai}: a technology that is directed at automating the end-to-end AI Lifecycle, from data cleaning, to algorithm selection, and to model deployment and monitoring in the ML workflow \citep{wang2020autoai}.
    \item AutoMLPipeline (AMLP)~\citep{amlp}: a toolbox that provides a semi-automated ML model generation and forecasting.
\end{itemize}

Models generated by the above AutoML technology were used to benchmark the predictive skill of the SPATIAL framework, provide additional insight into data characteristics and pattern extraction of environmental datasets, and assess the ease of deployment of the different model frameworks. 

\begin{figure*}[t]
\centering
    \includegraphics[width=0.85\textwidth]{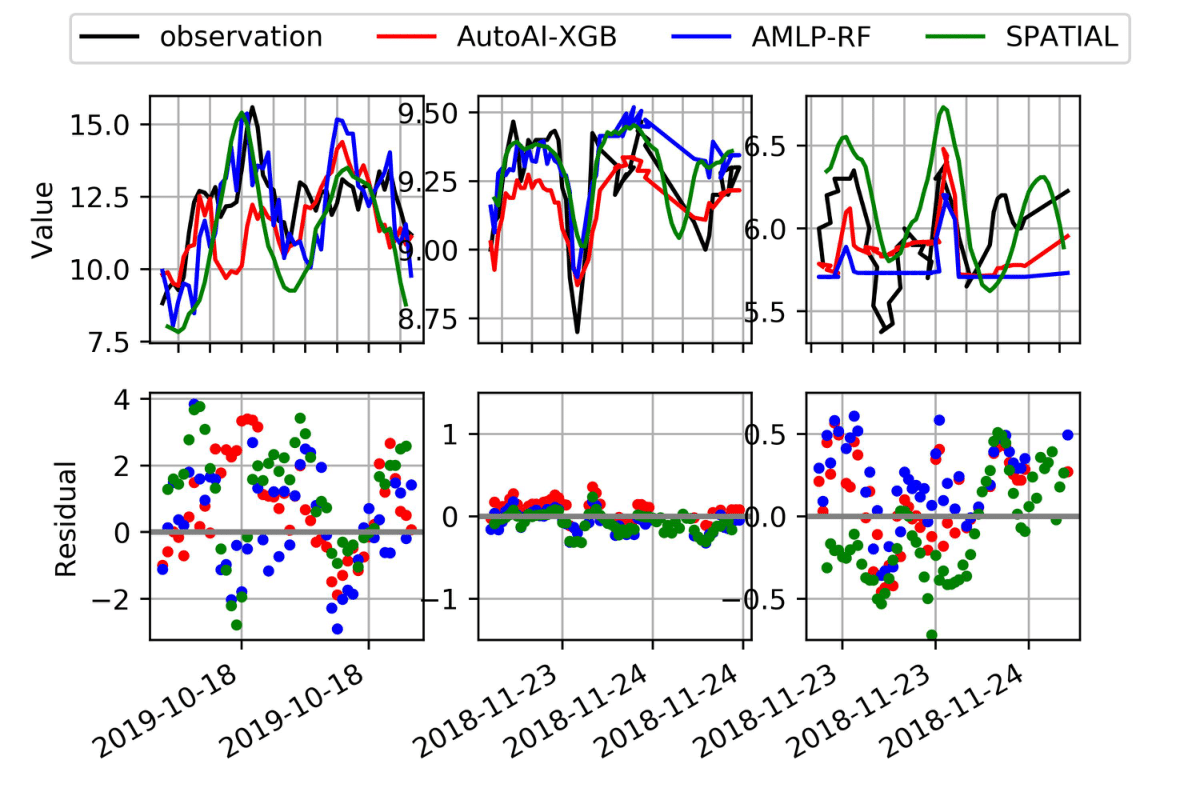}
    \caption{Comparison of SPATIAL performance against two benchmark models: An XGBoost model configured and deployed by AutoAI (red line), and a Random Forest model generated by using the AMLP pipeline (blue line). Black line indicates observations. The top row compares each model against observations while the bottom row presents the residual plots between observation and prediction. Each model predicts a 24-hour ahead forecast over 24 hours with a stride of 30 minutes (new forecast generated every 30 minutes). Data is presented for ADCP (left column), Dissolved Oxygen (middle column), and Temperature (right column).}
\label{fig:tsresults}
\end{figure*} 

\medskip

Figure~\ref{fig:allsensordata} presents all sensor data illustrating a number of characteristics:
\begin{itemize}
    \item The current speed is highly volatile and reports rapid variations over a relatively large range between 0 -- \SI{75}{cm/s}.
    \item Temperature data exhibits a pronounced seasonal pattern, reducing from a peak of about 18 $^{\circ}$C to 3 $^{\circ}$C in latter half of December.
    \item DO is exposed to relatively large fluctuations that are influenced by exogenous, non-seasonal factors (particularly aquaculture-driven impacts such as supplemental oxygenation and fish metabolism). Values drop rapidly from a peak of about \SI{10}{mg/L} in late August to a minimum of about \SI{7}{mg/L} in mid-October before increasing back to approximately \SI{10}{mg/L}. 
\end{itemize}

These pronounced patterns encapsulated a wide-ranging array of time-series behaviours including volatility, seasonality, and exogenous drivers. Of particular interest are seasonal patterns in temperature data and the capabilities to learn seasonally-varying patterns from short, sub-seasonal datasets (both temperature and DO contain 17 weeks of observation data).
These constitute a challenging time-series forecasting problem and early analysis of standard approaches such as GAM and ARIMA modeling reported substandard predictive skill. 

\begin{table}[htbp]
    \begin{center}
\begin{tabular}{| *{7}{c|} }
    \hline\hline 
    & \multicolumn{2}{c|}{SPATIAL}
            & \multicolumn{2}{c|}{AMLP-RF}
                    & \multicolumn{2}{c|}{AutoAI-XGB} \\
    \hline
    Dataset   & MAE        & $\sigma$ &   MAE & $\sigma$   &   MAE & $\sigma$    \\
    \hline
      Temp   & 0.86 & 0.54  &   0.25   &   0.04  &   0.21  &   0.03    \\
  %  \hline
        DO   &  0.32    & 0.25   &   0.27   &   0.04  &   0.28  &   0.04    \\
  %  \hline
      ADCP   &  4.67      &  5.40    &   3.80   &  1.68   &   4.51   &   1.94   \\
    \hline \hline
\end{tabular}
    \end{center}
        \caption{The MAE and standard deviation of reported MAE across all three datasets for a five-fold cross-validation interrogation. }
    \label{tab:results}
\end{table}

We compared our SPATIAL model against the two baseline implementations described previously. Tuning of model performance considered effects of appropriate number of time lags and hyperparameter selection, and adopted a 5-fold cross-validation for all models. The baseline models relied on a robust automated hyperparameter optimisation routine, while hyperparameter tuning of the SPATIAL adopted a greedy grid-search approach. 

Table \ref{tab:results} summarises model performance in terms of MAE and the standard deviation of the error across series of cross-validation tests. 
The baseline models that demonstrated best performance across all datasets were XGBoost and Random Forest, for AutoAI (AutoAI-XGB) and AMLP (AMLP-RF), respectively. 
We see strong performance of the different models across all datasets, despite the high complexity of the data. 
ADCP predictions reported significantly higher MAE for all models, which was largely a result of larger volatility and range in the data. Table \ref{tab:summarystat} presents summary statistics of the three datasets indicating that the standard deviation varies across the datasets. Considering the reported MAE relative to the standard deviation of the data, we see comparable predictive skill across the three datasets.

 Figure~\ref{fig:tsresults} provides a time series plot comparing the three models against observations (we selected one sensor from each of the dataset to illustrate). All models closely capture both the short-scale fluctuations and seasonal patterns of the data. This is particularly true for data such as temperature which has a pronounced seasonal pattern and can be challenging for ML models to learn with less than one-year of data (in our case, our training data was 17 weeks). 

Performance comparisons demonstrated that existing algorithms give strong predictive skill on these datasets. This performance is broadly replicated by our SPATIAL model. Namely, adopting a simpler data preprocessing, and model generation and implementation pipeline (fewer models to support by grouping sensor data), we get performance comparable to existing state-of-the-art. Our SPATIAL model has a number of practical advantages against existing algorithms, particularly:
\begin{enumerate}
    \item The data preprocessing pipeline is simplified as the SPATIAL pipeline loads all data into a single array that is fed to the networks (the model does not rely on exogenous variables or feature transformations). Missing data is handled readily with masking layer. This compares with the the benchmark models that requires a bespoke model for each sensor signal and transforming into appropriate matrix structure dependent on the desired prediction window, historic features and stride forward step length~\cite{tsml2020}.  
    \item By feeding data from multiple sensors that are geographically in proximity or share certain characteristics, the network can capture certain physical relationships that exist in nature and potentially serves to regularize the model.
    \item Treating all sensors simultaneously provides a large increase in computational efficiency -- only one model is trained rather than a model for each signal.
\end{enumerate}

Points 1 and 2 above are closely connected. In forecasting, we wish to use the simplest model that most closely mirrors what we know about the system. The proposed SPATIAL framework simplifies the model training and deployment process for environmental modeling by: 
\begin{itemize}
    \item requiring only one model be trained and maintained;
    \item eliminating need to implement transformations or matrix manipulations of the input data;
    \item  naturally generating a time series sequence for the desired forecasting period rather than a single value for each direct model (although classical approaches can adopt an iterative method to generate a time-sequence prediction via multi-step forecasting, predictive skill is generally limited \citep{hamzaccebi2009comparison}).
\end{itemize}

An important point is that the SPATIAL model processes information from multiple sensors that are known to be connected (based on the spatiotemporal nature of the dataset). There is a large volume of literature related to applying physics-informed constraints to ML which is discussed in the Related Work section. The objective in many of those studies is to augment the models with data external to the training data via aspects such as modified loss functions, data augmentation, or specifying consensus filters to guide disparate models or data towards convergence \citep{haehnel2020using}. The proposed methodology presents a natural framework to ingest information external to the time series signal to the predictive model positing the opportunity to enhance learning.
% The concept of using physics-based constraints and physics-informed loss functions is well established in the literature  and has demonstrated improved regularisation and robustness of machine learning models \citet{raissi2019physics}.
Within a time series forecasting paradigm, the nominal architectures that deliver performance are well-established, and often the challenge reduces to appropriate data conditioning and feature engineering. Missing data, anomalous data, and sensor drift are major obstacles, particularly for ocean-related deployments. SPATIAL presents a natural framework to reduce the impacts of individual sensor error or bias by enabling parameter sharing and pattern learning across multiple sensors that can serve to regularize individual deviations.

Further, we investigated how interdependence of different variables can inform learning. Dissolved oxygen has a nonlinear dependency on temperature and both data were sampled at the same location. In a classical sense, one could adopt temperature as a feature to a ML framework. We explored a combined oxygen and temperature SPATIAL that fed data from both simultaneously to the network. Results demonstrated reduced MAE compared to treating both independently indicating the ability to learn from multiple scales. The combined SPATIAL reported 
MAE of \SI{0.3}{mg/L}, 
and \SI{0.65}{^\circ C} for oxygen and temperature, respectively, when SPATIAL was applied across both datasets simultaneously, while the corresponding results for the individual models were 0.32 and 0.86 for oxygen and temperature. 

Ocean variables such as dissolved oxygen and temperature are intrinsically connected and it is intuitive to apply the capabilities of DL to extract and exploit these relationships to achieve uplift in predictive skill. Traditionally, this is a part of the feature engineering phase. SPATIAL however, provides a natural framework to readily explore these relationships via a simple DL framework that extracts these spatiotemporal patterns. 

Finally, this paper compares SPATIAL against two baseline frameworks. Importantly, the baseline models required training on each dataset individually, representing a total of 20 model training iterations compared to three for the SPATIAL. 
The computational saving is approximately similar (six-fold reduction), although in reality the savings are larger, as the SPATIAL architecture is relatively simple and required minimal hyperparameter tuning compared to a full algorithm search implemented by AutoML approaches. 
These computational savings are particularly attractive for edge-computing applications and is a major area of interest for marine industry applications \citep{ODonncha2020}. Many ocean sensor deployments consist of multiple sensors connected to a low-power compute device to manage storage, data processing, and connectivity. Extending these with a lightweight predictive framework that can forecast based only on the data collected from the sensors can allow for more intelligent data processing and enable adaptive sampling for sensor networks \citep{jain2004adaptive}.

\section{Conclusions}
\label{sec:conclusions}

This paper presents a DL framework based on bidirectional LSTM to extend learning of time series signals. The resulting model provides a scalable forecasting system that adjusts naturally to spatiotemporal patterns. The cost of model training was reduced by an order of magnitude as we trained a single model for each \textit{dataset} rather than each \textit{signal}. SPATIAL adjusts naturally to missing data through a simple masking layer and the random distribution of missing data between sensors can be leveraged to lessen the effect of data gaps on learning (i.e. since typically sensors exhibit data gaps at different times, combining information from multiple sensors in a single framework can improve learning on noisy and error-prone data). Finally extending the recurrent structure of LSTM-type approaches for time series applications to the spatiotemporal direction has potential to translate to many different industry applications such as weather, transport, and epidemiology.

The ML approach is extremely useful in forecasting oceanographic conditions of water quality (e.g. oxygen), especially as it assimilates realtime forcing variables such as temperature from sources including sensor networks or remote sensing. As additional big data layers become available from instruments such as underwater autonomous vehicles, this approach can be extended to 3D and 4D geospatial data to map nearshore dynamics of oxygen or other variables. These forecasts are essential in management decisions regarding coastal resources including aquaculture, fisheries, and pollution discharge.

%Value of the approach.
%Potential to integrate different models (from different orgs, etc.) (merging)
%To speed-up the method, one could consider methods from on-line and time-varying
%optimisation.
%\section{References}
\bibliography{library}
\end{document}